\documentclass[journal]{IEEEtran}
\usepackage[english]{babel}
\usepackage[utf8]{inputenc}

\usepackage{amsmath,amssymb,mathrsfs,bbm}
\usepackage{graphicx}
\usepackage{multirow,tabularx}
\usepackage[lined,linesnumbered,ruled]{algorithm2e}
\usepackage{url}

\usepackage[font=footnotesize]{caption}

\usepackage{psfrag,epsfig,graphics}
\usepackage{amsmath,amssymb,multirow}
\usepackage{mathbbol}
 
\usepackage{subcaption}
\usepackage{makecell}
\usepackage{footnote}
\usepackage{threeparttable}
\usepackage{booktabs, caption}
\usepackage{textcomp}
\usepackage{mathtools}
\usepackage{xcolor}
\usepackage{graphicx}              
\usepackage{array}      
\usepackage{siunitx}
\usepackage[normalem]{ulem}

\DeclareSymbolFontAlphabet{\amsmathbb}{AMSb}%
\newcommand{\cp}[1]{\ifmmode {\mathcal{#1}}\else ${\mathcal{#1}}$\fi}
\newcommand{\bA}{\mathbf{A}}

\newcommand{\bN}{\mathbf{N}}
\newcommand{\bP}{\mathbf{P}}

\newcommand{\bW}{\mathbf{W}}
\newcommand{\bX}{\mathbf{X}}
\newcommand{\bY}{\mathbf{Y}}
\newcommand{\bZ}{\mathbf{Z}}
\newcommand{\ba}{\mathbf{a}}
\newcommand{\bb}{\mathbf{b}}

\newcommand{\by}{\mathbf{y}}

\newcommand{\bx}{\mathbf{x}}

\newcommand{\bZhat}{\,\hat{\!\bZ}}

\newcommand{\bYhat}{\,\hat{\!\bY}}

\newcommand{\calG}{\mathcal{G}}

\newcommand{\bPi}{\mathbf{\Pi}}

\usepackage{color}  %

\def\cblue{\textcolor{black}}

\definecolor{darkgreen}{rgb}{0., 0., 0.}
\definecolor{dorange}{rgb}{0., 0., 0.0}
\newcommand{\cdgreen}{\textcolor{black}}

\newcommand{\cdorange}{\textcolor{black}}

\definecolor{atomictangerine}{rgb}{1.0, 0.6, 0.4} %

\newcommand\scalemath[2]{\scalebox{#1}{\mbox{\ensuremath{\displaystyle #2}}}}

\begin{document}

\title{A Generalized Multiscale Bundle-Based Hyperspectral Sparse Unmixing Algorithm}

\author{Luciano C. Ayres, Ricardo A. Borsoi, José C. M. Bermudez, Sérgio J. M. de Almeida
\thanks{L. C. Ayres, J. C. M. Bermudez, Universidade Federal de Santa Catarina, Florian\'opolis-SC, e-mail: lucayress@gmail.com, j.bermudez@ieee.org; R. A. Borsoi, Universit\'e de Lorraine, CNRS, CRAN, Vandoeuvre-l\`es-Nancy, e-mail: raborsoi@gmail.com; S. J. M. de Almeida, Universidade Católica de Pelotas, Pelotas-RS, e-mail: sergio.almeida@ucpel.edu.br. \\
\text{\ \ \ }This work has been supported in part by the National Council for Scientific and Technological Development (CNPq), Coordination of Superior Level Staff Improvement (CAPES) and Foundation for Research Support of the State of Rio Grande do Sul (FAPERGS).
}%
}

\maketitle

\begin{abstract}
	In hyperspectral sparse unmixing, a successful approach employs spectral bundles to address the variability of the endmembers in the spatial domain. However, the regularization penalties usually employed aggregate substantial computational complexity, and the solutions are very noise-sensitive. We generalize a multiscale spatial regularization approach to solve the unmixing problem by incorporating group sparsity-inducing mixed norms. Then, we propose a noise-robust method that can take advantage of the bundle structure to deal with endmember variability while ensuring inter- and intra-class sparsity in abundance estimation with reasonable computational cost. We also present a general heuristic to select the \emph{most representative} abundance estimation over multiple runs of the unmixing process, yielding a solution that is robust and highly reproducible. Experiments illustrate the robustness and consistency of the results when compared to related methods.
\end{abstract}


\section{Introduction}

In hyperspectral image (HSI) analysis, spectral unmixing (SU) consists of determining the spectral signatures of the materials contained in the scene (\emph{i.e.}, the \textit{endmembers} -- EMs) and the proportions in which they are present in each pixel of the image~\cite{bioucas2012hyperspectral}. Most SU approaches are based on a linear mixture model (LMM) \cite{bioucas2012hyperspectral}, which assumes that each pixel in the HSI consists of a linear combination of the EMs, weighted by their fractional abundances.
However, variations in atmospheric, lighting, and environmental conditions that commonly occur in a scene can have a significant impact on the spectral signatures of the EMs contained in an HSI \cite{borsoi2021spectral}. Because of this \emph{spectral variability}, the spectrum representing an EM can change as a function of its position in the HSI.

The use of large libraries of spectra is a typical approach to deal with spectral variability in SU \cite{borsoi2021spectral}. Among existing methods, those based on sparse regression, considered computationally efficient, assume that the reflectance of pixels in an HSI can be described as a linear combination of a few EM signatures from a large spectral library known \emph{a priori} (typically constructed from laboratory or \emph{in situ} measurements) \cite{iordache2011sparse}.
However, spectral libraries are often not available for a given HSI. Recent work proposed methods for extracting structured spectral libraries directly from the observed HSI~\cite{somers2012automated}. Such approaches usually apply EM extraction algorithms (EEAs) to subsets of pixels randomly sampled from the HSI. This randomness causes EMs obtained at each extraction to be slightly different, representing the spectral \mbox{variability in the HSI.}

In \cite{drumetz2019hyperspectral}, the authors propose to introduce mixed norms in the sparse SU optimization problem with structured spectral libraries to promote group sparsity. A new penalty is proposed to control inter- and intra-structure sparsity, which can considerably improve SU performance when compared to typically used sparsity penalties.
However, the results of this technique depend on the initialization of the abundances in the optimization problem. Furthermore, when used in conjunction with spectral libraries extracted from the HSI via methods such as~\cite{somers2012automated}, the estimated abundances are random variables and may be different for each run of the method.

Another challenge with these techniques is their sensitivity to the presence of noise due to the large number of signatures in the library. Integrating regularizations that promote spatial smoothness of the abundances improves the performance of sparse SU algorithms in noisy conditions, but at the expense of a considerable increase in computational complexity~\cite{iordache2012total, zhang2018spectral}. A fast sparse SU algorithm (called MUA), based on superpixels and a multiscale strategy, has recently been proposed for the SU problem without using structured libraries~\cite{borsoi2018fast}. Despite the positive results obtained in~\cite{borsoi2018fast,ayres2021homogeneityICASSP} with a sparsity penalty based on the L$_1$ norm, this solution is not suitable to tackle the challenge of spectral variability through structured libraries.

This work proposes a method to handle spectral variability of EMs in the SU problem while maintaining a reasonable computational cost and being robust to noise. Motivated by methods used in neuroimaging~\cite{du2014novel}, we aim to obtain \emph{reproducible} SU solutions~\cite{adali2022reproducibility}, i.e., using the same algorithm and data, the results should be consistent, even considering randomness in some steps of the method.
The main contributions of this paper can be summarized as: 1) We generalize the multiscale spatial regularization problem formulated in the MUA algorithm \cite{borsoi2018fast} to solve the sparse SU problem with structured libraries, allowing the use of more general sparsity-inducing penalties, such as those based on mixed norms~\cite{drumetz2019hyperspectral}; 2) We propose a heuristic %
based on a graph centrality criterion to select reproducible abundance estimates (i.e., with less influence of randomness on the SU process) from $K$ runs of the method.
Experimental results show that the proposed method provides high quality results with significantly less dispersion when compared to state-of-the-art algorithms. 
Codes to reproduce the experimental results are available at \begin{color}{blue}\url{http://github.com/lucayress/GMBUA}\end{color}.

\section{Structured sparse spectral unmixing} \label{sec:bundles_norms}

Considering the LMM, an HSI $\bY \in \amsmathbb{R}^{L \times N}$ with $L$ bands, $N$ pixels and $P$ EMs can be written as:
\begin{equation}
    \bY = \bA\bZ+\bN \,,
    \label{eq:LMM}
\end{equation}
where the abundances $\bZ \in \amsmathbb{R}^{P \times N}$ are subject to the non-negativity (ANC) and sum-to-one (ASC) constraints \cite{bioucas2012hyperspectral}. $\bA \in \amsmathbb{R}^{L \times P}$ represents the matrix of EMs. Column $\mathbf{a}_p$, $p=1,\dots,P$, of $\bA$ is the signature of of the $p$th EM. $\bN$ is the additive noise. Using LMM allowing for consideration of spectral variability requires a representation of any given material using more than one EM.

\subsection{SU with structured EM libraries}
Following \cite{drumetz2019hyperspectral}, consider replacing $\bA$ in LMM~\eqref{eq:LMM} with a structured spectral library $\mathbf{B} \in \amsmathbb{R}^{L \times Q}$. $\mathbf{B}$ is composed of $P$ structures $\mathbf{B}_p \ (p=1,\dots,P)$. The columns of the $p$th structure are different spectral signatures of the $p$th EM of the HSI:
\begin{equation}
    \bY = \mathbf{BX}+\bN \,, \qquad \mathbf{B} = \big[\mathbf{B}_1 \ |\ \mathbf{B}_2 \ |\ \dots \ |\ \mathbf{B}_{P}\big] \,.
    \label{eq:LMM_bundle}
\end{equation}
In \eqref{eq:LMM_bundle}, $\bX \in \amsmathbb{R}^{Q \times N}$ contains the abundances associated with each spectral signature in $\mathbf{B}$. Each submatrix $\mathbf{B}_p \in \amsmathbb{R}^{L \times m_{\calG_p}}$ represents a group $\calG_p$ of $m_{\calG_p}$ signatures. Then, $Q = \sum_{p=1}^P m_{\calG_p}$ is the total number of spectra. The $i$th representative of the $\calG_p$ group in the dictionary is denoted as $\bb_{{\calG_p},i}$ (i.e., $\bb_{{\calG_p},i}$ is the $i$th column of $\mathbf{B}_p$). %
We can extract from \eqref{eq:LMM_bundle} the expression of a single pixel $\by_n$, $n=1,\dots,N$, in terms of the ``global'' abundance of the $p$th material~\cite{drumetz2019hyperspectral}:
\begin{equation}
    \by_n = \sum_{p=1}^{P} z_{p}\ba_p = \sum_{p=1}^{P}\Biggl( \sum_{i=1}^{m_{\calG_p}} x_{{\calG_p},i}\Biggl)\Biggl(  \frac{\sum_{i=1}^{m_{\calG_p}} x_{{\calG_p},i}\bb_{{\calG_p},i}}{\sum_{i=1}^{m_{\calG_p}} x_{{\calG_p},i}} \Biggl).
    \label{eq:bundle_pxl_global}
\end{equation}
where $x_{{\calG_p},i}$ is the abundance associated with $\bb_{{\calG_p},i}$, and $z_{p}$ is the total abundance coefficient for material $p$ in $\by_n$.

Sparse SU consists of recovering the $\bX$ abundances from the observed $\bY$ HSI, given the $\mathbf{B}$ dictionary. We can then formulate the following optimization problem:
\begin{align}
    \hat{\bX} = \underset{\bX\in\Delta_P}{\operatorname{arg\ min}}\ \frac{1}{2}\|\bY-\mathbf{B}\bX\|^2_F + \lambda\mathcal{R}(\bX) \,,
    \label{eq:traditional_sparse_SU}
\end{align}
where $\mathcal{R}(\bX)$ is a regularization penalty that promotes sparsity, $\lambda\in\amsmathbb{R}_+$ is a regularization parameter, $\|\cdot\|_F$ is the Frobenius norm, and $\Delta_P$ denotes the set of coefficient matrices whose abundances satisfy the ANC and ASC constraints.

\subsection{Mixed sparsity-promoting norms (penalties)}

The regularization penalty $\mathcal{R}(\bX)$ plays a key role in the performance of sparse SU algorithms.
In \cite{drumetz2019hyperspectral}, Drumetz \emph{et al.} propose the use of group sparsity by introducing mixed norms in the sparse SU optimization problem to enforce sparsity within each group and between different groups from structured libraries of EMs automatically extracted from HSI as in \cite{somers2012automated}.
The mixed two-level norm $\ell_{\calG,r,s}$ is defined for any pair of positive real numbers, $r$ and $s$, as:
\begin{equation}
    \|\bx\|_{\calG,r,s} \triangleq \Biggl( \sum_{p=1}^{P} \Biggl( \sum_{i=1}^{m_{\calG_p}} |x_{\calG_{p,i}}|^r \Biggl)^{s/r}\Biggl)^{1/s} = \Biggl( \sum_{p=1}^{P} \|\bx_{\calG_p}\|_{r}^{s} \Biggl)^{1/s} \!.
    \label{eq:mix_norm_vec}
\end{equation}
Operating on columns and summing the results across all pixels, the authors in~\cite{drumetz2019hyperspectral} defined the sparsity penalty through the $\ell_{\calG,r,s}$ norm of the coefficient matrix as 
$\mathcal{R}(\bX) =\|\bX\|_{\calG,r,s} \triangleq \sum_{n=1}^{N} \|\bx_{n}\|_{\calG,r,s}$.
Considering this penalty, the sparse SU problem~\eqref{eq:traditional_sparse_SU} was solved in~\cite{drumetz2019hyperspectral} using the alternating direction method of multipliers. However, $\mathcal{R}(\bX)$ is non-convex for various values of $r$ and $s$, which makes the solution computationally costly and initialization dependent. 
We highlight that while extracting the $\mathbf{B}$ library from the HSI \cite{somers2012automated} can relieve the need for spectral libraries to be known a priori, this strategy has limitations as it relies on the presence of pure pixels and on the EEAs yielding reliable results. Moreover, it can add significant variability to the SU results.

\section{Proposed method} \label{sec:method}

In this section, we present the proposed structured sparse SU algorithm (called GMBUA -- \emph{Generalized Multiscale Bundle-based Unmixing Algorithm}), which has two key parts. First, we develop a multiscale spatial regularization approach capable of considering a wide variety of sparsity penalties, such as the norm $\ell_{\calG,r,s}$, in an efficient manner. Then, we propose a strategy to mitigate the effect of randomness involved in extracting the spectral library from the HSI (as in, e.g.,~\cite{somers2012automated}) and in the SU process. This strategy uses a centrality criterion applied to a graph representing different abundance realizations.

\subsection{Multiscale approach for sparse and structured SU} \label{sec:gen_MUA}

Consider a spectral library $\mathbf{B}$ extracted from HSI using the an algorithm such as, e.g., \cite{somers2012automated}. We define the multiscale decomposition of the sparse SU problem at~\eqref{eq:traditional_sparse_SU} as a spatial transformation promoted by an operator $\mathbf{W} \in \amsmathbb{R}^{N \times M}$, $M<N$, applied to the HSI and to the abundances~\cite{borsoi2018fast}:
\begin{equation} \label{eq:sparse_Yc}
	\bY_{\!\mathcal{C}} = \mathbf{YW}, \qquad \bX_\mathcal{C} = \mathbf{XW},
\end{equation}
where $\bY_{\!\mathcal{C}} \in \amsmathbb{R}^{L \times M}$, $\bX_\mathcal{C} \in \amsmathbb{R}^{Q \times M}$, and subscript $\mathcal{C}$ refers to the new approximate (coarse) image domain. The $\bW$ operator, built based on a superpixel decomposition of $\bY$, averages the pixels in spatially homogeneous regions of the HSI where the abundances are relatively constant. This spatial smoothness of the abundances can then be exploited by formulating the sparse SU problem in the approximate domain according to~\eqref{eq:sparse_Yc}:
\begin{equation}\label{eq:opt_coarse_Xc}
	\hat{\bX}_\mathcal{C} = \underset{\bX_{\mathcal{C}} \in\Delta_P}{\operatorname{arg\ min}}\ \frac{1}{2}\|\bY_{\!\mathcal{C}}-\mathbf{B}\bX_\mathcal{C}\|^2_F + \lambda_\mathcal{C}\mathcal{R}(\bX_\mathcal{C}) \,.
\end{equation}
Note that in \eqref{eq:opt_coarse_Xc} $\mathcal{R}$ is applied to the abundances at the approximation scale. Hence, as long as $\mathcal{R}$ does not directly use the spatial organization of the different pixels in the image, this procedure can be extended to any sparsity penalty. %
Problem \eqref{eq:opt_coarse_Xc} can be solved in the same way as \eqref{eq:traditional_sparse_SU}.
The $\hat{\bX}_\mathcal{C}$ estimate is then mapped back to the original spatial scale, denoted by $\mathcal{D}$, via the operation $\hat{\mathbf{{X}}}_\mathcal{D} = \hat{\mathbf{{X}}}_\mathcal{C}\mathbf{W}^* \in \amsmathbb{R}^{Q \times N}$, where operator $\mathbf{W}^*$ replicates the pixels from $\hat{\mathbf{{X}}}_\mathcal{C}$ at all pixel locations of the same superpixel at the $\mathcal{D}$ scale. Thus, $\hat{\mathbf{{X}}}_\mathcal{D}$ provides an estimate of the spatial structure of the abundance maps.

Next, the approximate $\hat{\mathbf{{X}}}_\mathcal{D}$ abundance matrix is used to regularize the SU problem at the original scale:
\begin{equation}\label{eq:opt_original}
	\hat{\bX} = \underset{\bX \in\Delta_P}{\operatorname{arg\ min}}\ \frac{1}{2}\|\bY-\mathbf{B}\bX\|^2_F + \lambda\mathcal{R}(\bX) + \frac{\beta}{2}\|\hat{\mathbf{{X}}}_\mathcal{D}-{\mathbf{{X}}} \|^2_F,
\end{equation}
where $\beta$ is a regularization parameter.
The higher complexity of the solutions of \eqref{eq:opt_coarse_Xc} and \eqref{eq:opt_original} has been efficiently solved in~\cite{borsoi2018fast} for the case where $\mathcal{R}(\bX)=\|\bX\|_{1,1}$. This solution, however, cannot easily be generalized to other norms.

To enable simple solutions for other norms, we reformulate the multiscale approach by expressing \eqref{eq:opt_original} equivalently as:
\begin{align}\label{eq:opt_original2}
	\hat{\bX} = \underset{\bX \in\Delta_P}{\operatorname{arg\ min}}\ \frac{1}{2}\bigg\|
	\begin{bmatrix}
		\bY \\ \sqrt{\beta} \hat{\mathbf{{X}}}_\mathcal{D}
	\end{bmatrix} 
	- \begin{bmatrix}
		\mathbf{B} \\ \sqrt{\beta}\mathbf{I}
	\end{bmatrix} 
	\bX\bigg\|^2_F
	+ \lambda\mathcal{R}(\bX).
\end{align}
By defining $\widetilde{\bY}=\big[\bY^\top,\sqrt{\beta}\hat{\mathbf{{X}}}_\mathcal{D}^\top\big]^\top$ and $\widetilde{\mathbf{B}}=\big[\mathbf{B}^\top,\sqrt{\beta}\mathbf{I}\big]^\top$, this problem can be written in the same form as~\eqref{eq:traditional_sparse_SU}:
\begin{equation}\label{eq:opt_original2_bundle}
    \hat{\bX} = \underset{\bX \in\Delta_P}{\operatorname{arg\ min}}\ \scalemath{0.8}{\frac{1}{2}} \big\|\widetilde{\bY}-\widetilde{\mathbf{B}}\bX\big\|^2_F + \lambda\mathcal{R}(\bX) \,.
\end{equation}
Problem \eqref{eq:opt_original2_bundle} is equivalent to the sparse SU \eqref{eq:traditional_sparse_SU} without spatial regularization, and with an HSI and dictionary given by $\widetilde{\bY}$ and $\widetilde{\mathbf{B}}$, respectively. Consequently, both~\eqref{eq:opt_coarse_Xc} and~\eqref{eq:opt_original2_bundle} can be solved using the same algorithms developed to solve the problem~\eqref{eq:traditional_sparse_SU}. In particular, in this paper we will consider $\mathcal{R}(\bX)$ to be the mixed norm $\ell_{\calG,r,s}$. %
This allows us to use the strategy described in~\cite{drumetz2019hyperspectral} to solve the optimization problems. We omit the details for brevity.
After solving~\eqref{eq:opt_original2_bundle}, the global abundances $\hat{\bZ}$ are then calculated by aggregating the abundances in each group of solution $\hat{\bX}$ (as shown in~\eqref{eq:bundle_pxl_global}).

\subsection{Selecting a consistent abundance estimate}

The estimated abundances $\hat{\mathbf{X}}$ are random variables due to randomness in both the spectral library extraction and SU processes. 
Since library extraction typically involves applying EEAs to randomly sampled pixels of the HSI significant variability can occur between different executions of the algorithm, even when the EEA is reliable.
This undesirable variability has been often ignored, and choosing a representative solution is not a trivial task. Inspired by strategies recently considered for neuroscience applications~\cite{du2014novel}, we propose a strategy to obtain a representative and reproducible estimate.

First, assuming there exist pure pixels of each EM in the image, $K$ different spectral libraries are extracted from HSI $\bY$ using a method such as, e.g.,~\cite{somers2012automated}. For each of them, SU is performed using the algorithm described in the previous subsection, leading to $K$ different abundance maps $\hat{\bZ}_k$ with $(k=1,\dots,K)$, which are stored in the set $\Omega = \bigl\{\hat{\bZ}_1, \dots, \hat{\bZ}_K\bigl\}$. Now, we construct a graph with $K$ nodes, each representing one of the solutions in $\Omega$. The weighted adjacency matrix of the graph is constructed based on a similarity measure $C_{uv}$ between each pair of solutions, $\hat{\bZ}_u$ and $\hat{\bZ}_v$, computed based on a linear assignment problem (LAP). More precisely, the similarity between $\hat{\bZ}_u$ and $\hat{\bZ}_v$, $u\neq v$ is calculated as
	\begin{align}
		C_{uv} = \min_{\bP\in\bPi} \,\,\scalemath{0.8}{\frac{1}{N}} \big\|\hat{\bZ}_u-\bP\hat{\bZ}_v\big\|_F \,,
		\label{eq:assignment1}
	\end{align}
where $\bPi$ represents the set of permutation matrices of dimension $P$. The minimization in~\eqref{eq:assignment1} aims to align the abundance maps in such a way as to eliminate permutation ambiguity in the estimated abundances, and is efficiently solved in the form of an LAP (using, e.g., the Hungarian algorithm \cite{kuhn1955hungarian}).

By calculating $C_{uv}$ for $u,v=1,\ldots,K$, $u\neq v$, we create a connected, undirected graph with no self-loops. A minimum-cost sub-graph connecting all nodes is the next step in computing a minimum spanning tree (MST)~\cite{kruskal1956shortest}, eliminating connections between unnecessary nodes. Finally, we select the most representative run as the most central node in the graph; this consists of the run corresponding to the node with the largest number of connections to other nodes (runs) in the graph. This solution, called the most representative, or the \emph{best}, is denoted by $\bZ_*$.

Note that the strategy proposed in this section is general and can be applied to select a representative solution for any SU algorithm that contains a significant degree of randomness.
A pseudocode for GMBUA is presented in Algorithm~\ref{alg:method}.

\vspace{-0.2cm}

\begin{algorithm}[h!] 
\footnotesize
\SetAlgoLined
\KwIn{$\bY$, number of runs $K$, of EMs $P$, and of EM extractions $T$, percentage of pixels $\alpha$, number of superpixels $M$, regularization parameters $\beta$, $\lambda$, $\lambda_\mathcal{C}$.}
$\Omega \leftarrow \emptyset $ (empty set); \\
Construct $\bW$ based on the superpixel decomposition of $\bY$; \\
    \For{$k=1$ \KwTo $K$}{ 
        $\Upsilon \leftarrow \emptyset $ (empty set); \\
        \For{$i=1$ \KwTo $T$}{
            $\Psi\leftarrow$ extract $P$ EM candidates from $\lceil\alpha N\rceil$ pixels randomly sampled from $\bY$; \\
            $\Upsilon \leftarrow \Upsilon \bigcup \{\Psi\}$ --- store EM candidates in $\Upsilon$; \\
        }
        $\{\mathbf{B}_i\}_{i=1}^P \leftarrow$ cluster the signatures in $\Upsilon$ into $P$ groups; \\
        $\hat{\bX} \leftarrow$ perform multiscale unmixing solving~\eqref{eq:opt_coarse_Xc} and~\eqref{eq:opt_original2}; \\
        $\hat{\bZ}_k \leftarrow$ convert $\hat{\bX}$ into global abundances with $P$ EMs; \\
        $\Omega \leftarrow \Omega \bigcup \{\hat{\bZ}_k\}$ --- store each solution in $\Omega$;
    }
 
 \Return{$\hat{\bZ}_* \leftarrow$ \rm{the most representative abundance in the set} $\Omega$}.
 \caption{GMBUA} \label{alg:method}

\end{algorithm}

\vspace{-0.5cm}

\section{Experimental Results and Discussion} \label{sec:results}

In this section, we compare the GMBUA method with the constrained least squares method (FCLSU), with the \emph{Collaborative} \cite{iordache2013collaborative}, \emph{Group} \cite{meier2008group}, \emph{Elitist}, \cite{kowalski2009sparsity} and \emph{Fractional} sparse SU based on different penalties~\cite{drumetz2019hyperspectral}, and with  SUnCNN~\cite{rasti2021suncnn}, which is a deep learning-based method.
For quantitative comparisons, in addition to runtime, we use the signal-to-reconstruction error ratio in dB \cite{iordache2012total}, $\text{SRE}(\bZhat) = 10\log_{10}(\|\bZ\|^2_F/\|\bZ-\hat{\bZ}\|^2_F)$, which evaluates the estimation of the global abundances in cases where a reference abundance matrix (i.e., \emph{ground-truth} -- GT) is known. Also, we use $\text{SRE}(\bYhat) = 10\log_{10}(\|\bY\|^2_F/\|\bY-\hat{\bY}\|^2_F)$ to assess the reconstructed image $\bYhat$.
Qualitative evaluations were based on the visualization of the global abundance maps generated by the algorithms. Results were obtained for optimal regularization parameter values computed by grid search for each experiment and algorithm.
For the proposed method, we considered $K = 30$ runs to compute the most representative abundance maps in Algorithm \ref{alg:method}.
The algorithms were run in \emph{MATLAB$^{\text{\tiny \textregistered}}$}, on a computer equipped with a \emph{Intel Core i7 9750H} processor @ $2.60$ GHz and 16 GB of RAM.

\begin{figure}[t]
\centering
\centerline{\includegraphics[width=0.5\textwidth]{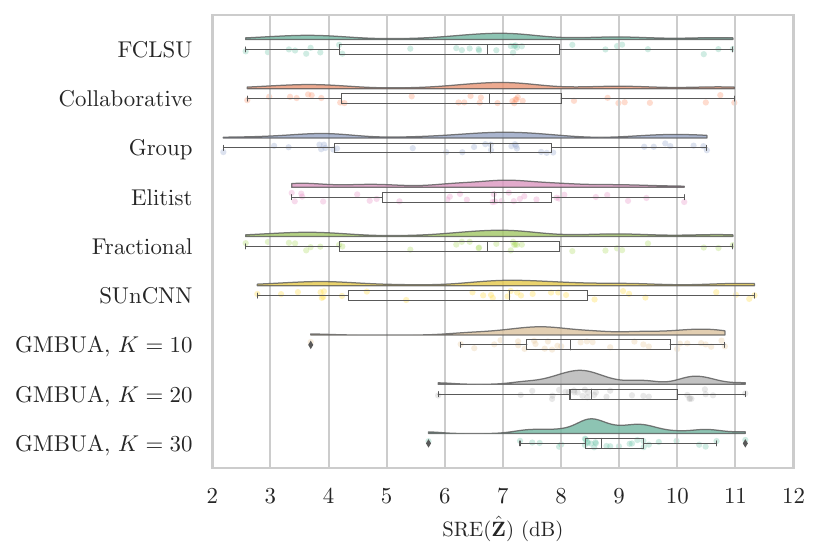}}
\vspace{-0.3cm}
\caption{SRE($\bZhat$) distribution for the synthetic data and all algorithms.}
\label{fig:synthetic4_dist}
\vspace{-0.5cm}
\end{figure}

\paragraph*{\textbf{Synthetic data (SD)}}

To quantitatively evaluate the performance of the algorithms, we generated an HSI from synthetically generated abundance maps with $50 \times 50$ pixels and $P=5$ EMs, which are used as GT. Less than 1.5\% of the pixels are \emph{pure} (i.e., have abundance greater than 0.95 for some EM).
To generate the EMs, a set of $P$ signatures were first extracted from the USGS library. These signatures were then multiplied by random piecewise linear functions as defined in~\cite{thouvenin2015hyperspectral} to generate the EM signatures used in each HSI pixel, incorporating spectral variability.
The pixels were generated following the LMM, where white Gaussian noise was added to the data to yield signal-to-noise ratios (SNR) of $20$ and $30$ dB.
To assess the reliability of the results generated by the algorithms, $R = 30$ Monte Carlo executions were performed. 
For each of the estimated abundances, we calculated the SRE($\bZhat$). The distributions of the respective results for each method are shown in Figure~\ref{fig:synthetic4_dist} (only shown for the case of 20 dB SNR due to space limitations). 
In order to show the influence of $K$ on the performance of GMBUA, we also show distribution results for $K = 10$ and $K=20$. It is clear from these results that GMBUA has led to a higher median SRE and to a significantly smaller spread of solutions than any of the other methods. Moreover, the spread of the distribution of SRE($\bZhat$) values decreases considerably with the increase of $K$,
significantly reducing the probability of a low-quality solution. These distributions indicate that GMBUA is robust to outliers, as it was able to generate consistently good results, with the majority of samples having SREs larger than $7$ dB. By contrast, the SRE($\bZhat$) values obtained using the competing methods varied widely within a range of approximately $2$ to $12$ dB, with a considerable amount of samples having SREs smaller than $5.5$ dB.

Table~\ref{tab:compare_synthetic} shows the median values of the SRE results. It can be verified from Table~\ref{tab:compare_synthetic} that the SUnCNN provided the highest median SRE($\bZhat$) for a 30 dB SNR, followed closely by the GMBUA, \emph{Fractional}, \emph{Collaborative} and FCLSU methods. Moreover, GMBUA performed significantly better than all the other algorithms for a 20 dB and 10 dB SNR, with gains of more than $1.5$ dB and $2.6$ dB in SRE($\bZhat$), respectively. The results of SUnCNN, on the other hand, were more sensitive to the increase in the amount of noise in the HSI.

The main cause for this variation in SRE values is related to the random sampling of image pixels during the construction of the spectral library from the HSI~\cite{somers2012automated}. This sampling leads to significant variations in the quality of the obtained libraries across runs, strongly impacting the results of the subsequent unmixing process, as seen in Figure~\ref{fig:synthetic4_dist}. Through the strategy for selecting the most consistent abundance run, GMBUA mitigates the impact of poor quality library extractions, increasing the reproducibility of the results.

\begin{table}[!t]

\renewcommand{\arraystretch}{1.1}
\fontsize{7.25pt}{7.25pt}\selectfont

\centering
      \caption{Quantitative results with the synthetic data (SD) and real data (Cuprite HSI) for all algorithms.}
      \vspace{-0.14cm}
      \centering
        \begin{tabular}{c||c|c|c|c} 
            \hline
            Method & Dataset & SRE($\bZhat$) & SRE($\bYhat$) & Runtime (s) \\
            \hline
            \multirow{4}{*}{FCLSU} & SD 30 dB & 10.58 & 28.04 & 1.02 \\
                                   & SD 20 dB & 6.58 & 29.93 & 1.01 \\\
                                   & SD 10 dB & 4.08 & 24.26 & 0.80 \\
                                   & Cuprite HSI & -- & 39.49 & 48.28 \\
            \hline
            \multirow{4}{*}{Collaborative} & SD 30 dB & 10.62 & 28.05 & 6.25 \\
                                   & SD 20 dB & 6.62 & 29.95 & 5.97 \\
                                   & SD 10 dB & 4.08 & 24.27 & 4.41 \\
                                   & Cuprite HSI & -- & 37.02 & 137.67 \\
            \hline
            \multirow{4}{*}{Group} & SD 30 dB & 9.88 & 26.09 & 8.94 \\
                                   & SD 20 dB & 6.79 & 29.86 & 9.95 \\
                                   & SD 10 dB & 4.11 & 24.72 & 4.28 \\
                                   & Cuprite HSI & -- & 33.74 & 273.42 \\
            \hline
            \multirow{4}{*}{Elitist} & SD 30 dB & 9.54 & 26.35 & 10.78 \\
                                   & SD 20 dB & 6.84 & 29.92 & 10.59 \\
                                   & SD 10 dB & 4.21 & 24.14 & 4.39 \\
                                   & Cuprite HSI & -- & 36.01 & 275.37 \\
            \hline
            \multirow{4}{*}{Fractional} & SD 30 dB & 10.59 & 28.06 & 8.90 \\
                                   & SD 20 dB & 6.59 & 29.94 & 8.93 \\
                                   & SD 10 dB & 4.08 & 24.26 & 6.75 \\
                                   & Cuprite HSI & -- & 36.97 & 278.59 \\
            \hline
            \multirow{4}{*}{SUnCNN} & SD 30 dB & 10.89 & 27.10 & 168.40 \\
                                   & SD 20 dB & 7.08 & 29.87 & 161.21 \\
                                   & SD 10 dB & 3.99 & 23.92 & 157.68 \\
                                   & Cuprite HSI & -- & 38.62 & 1142.99 \\
            \hline
            \multirow{4}{*}{GMBUA} & SD 30 dB & 10.72 & 26.31 & $K \times$17.04 \\
                                   & SD 20 dB & 8.60 & 28.82 & $K \times$17.70 \\
                                   & SD 10 dB & 6.79 & 25.12 & $K \times$13.77 \\
                                   & Cuprite HSI & -- & 37.06 & $K \times$344.98 \\
            \hline
        \end{tabular}
        \label{tab:compare_synthetic}
\end{table}

\begin{figure}[t]
\centering
\centerline{\includegraphics[width=0.9\linewidth]{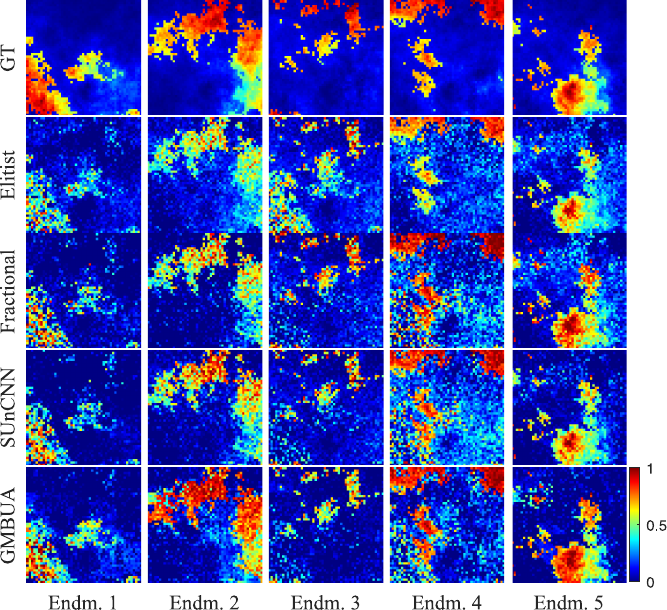}}
\caption{Estimated abundance maps for the synthetic dataset with an SNR of 20 dB. The sampled abundance maps are the ones corresponding to the median SRE($\bZhat$) results for each algorithm.}
\label{fig:synthetic_abund_maps}

\end{figure}

From Table~\ref{tab:compare_synthetic}, the \emph{Collaborative} and \emph{Fractional} algorithms obtained the highest SRE($\bYhat$) \cdorange{for the \cblue{cases} of 20 and 30 dB SNR}, providing the closest reconstruction of the HSI\cdorange{, while GMBUA and Group performed better in 10 dB SNR}. \cdgreen{However, a higher SRE($\bYhat$) does not imply a better abundance reconstruction performance since the unmixing problem is typically ill-posed, and different abundance solutions might yield similar SRE($\bYhat$).}
Observing the estimated global abundance maps for the \cdorange{$20$ dB SNR scenario} shown in Fig.~\ref{fig:synthetic_abund_maps} (the abundance corresponding to the FCLSU, \emph{Collaborative} and \emph{Group} algorithms were omitted due to space limitations), we can see that GMBUA generates abundances that are more similar to the reference one (GT), especially for the abundance maps of EMs 2 and 5. For the other EMs, the proposed method yields similar but less noisy abundance maps.

\begin{figure}[!t]
\centering
\centerline{\includegraphics[width=\linewidth]{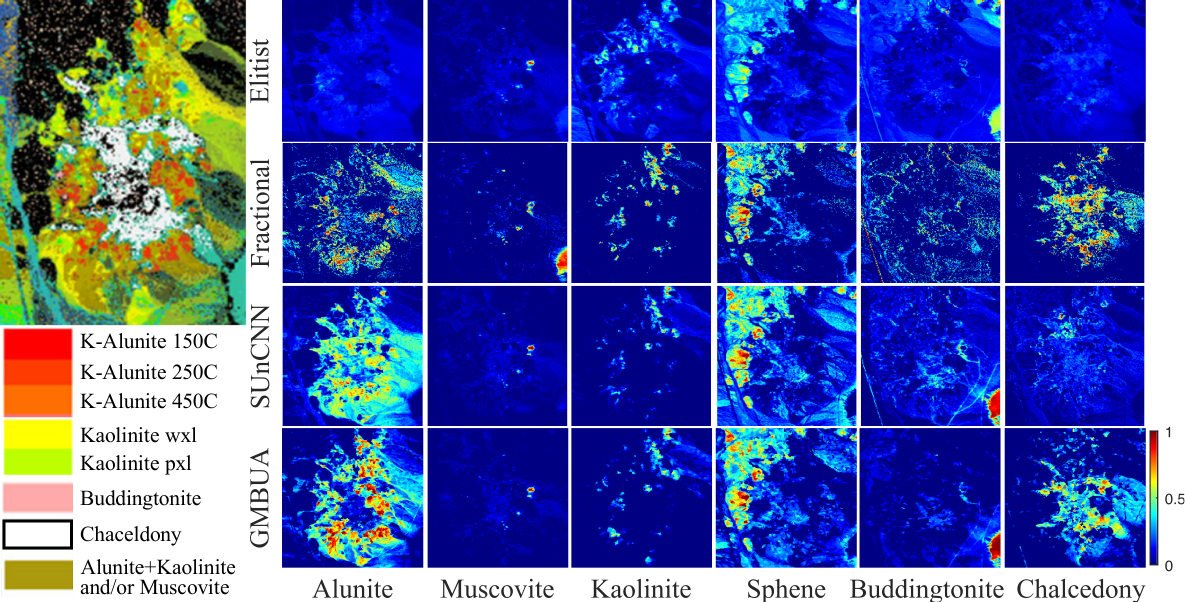}}
\caption{Estimated abundance maps for a random execution of the SU algorithms on the Cuprite HSI \cdorange{(a reference map is shown on the left).}}
\label{fig:real_abund_maps}
\end{figure}

\paragraph*{\textbf{Real data}} 
In this experiment, we used the \textit{well-known Cuprite} HSI~\cite{drumetz2019hyperspectral}, with $250 \times 191$ pixels and $188$ spectral bands. This HSI contains several exposed minerals, including alunite, buddingtonite, chalcedony, kaolinite, muscovite and sphene. 
Since we do not have access to a GT for this dataset, we rely on a qualitative evaluation of the estimated abundance maps.
Thus, we display in Figure~\ref{fig:real_abund_maps} the abundance maps obtained from a single execution of the algorithms, selected at random (the abundances corresponding to the FCLSU, \emph{Collaborative} and \emph{Group} algorithms are not displayed due to space limitations) \cdorange{and a reference map of different minerals in Cuprite scene for comparison.}

By observing the abundance maps, we see that the deep learning-based SUnCNN method in general could reasonably identify the materials in the scene but had some issues with alunite and chalcedony, which are not well separated. 
The \emph{Elitist} method did not perform well for this HSI, where the structures of kaolinite, sphene, and buddingtonite in the scene can be identified but are extremely mixed compared to the other methods.
\emph{Fractional} provided the noisiest abundances, and there was a considerable mixture between the muscovite and buddingtonite materials.
\mbox{GMBUA} provided the accurate and sparse abundance maps, presenting homogeneous regions with adequately separated components, most notably for the alunite mineral.

The SRE($\bYhat$) results and the execution times for all algorithms can be seen in Table~\ref{tab:compare_synthetic}. The FCLSU and SUnCNN methods achieved the highest SRE($\bYhat$), reconstructing the pixels very closely, whereas \emph{Group} achieved the lowest SRE($\bYhat$).
The execution times show that each of the $K$ unmixing runs inside GMBUA is comparable to the \emph{Fractional} method and much lower than those of SUnCNN. Thus, the proposed spatial regularization strategy does not add a considerable computation burden to the algorithm, with the extra time coming from the different executions required for selecting a consistent solution.

\section{Conclusions} \label{sec:conclusions}

In this paper, we proposed a generalized multiscale spatial regularization approach to solve the sparse SU problem with structured spectral libraries. The proposed method addresses the variability of EMs with robustness to noise while maintaining a reasonable computational complexity.
In addition, we proposed a graph-based approach for determining the most representative abundance estimate over multiple SU runs, significantly increasing the robustness to the randomness of the EMs extraction process and, thus, the reproducibility of the results.
Experiments demonstrated the superior performance of the proposed method when compared to related algorithms.

\vspace{-0.3cm}
\bibliographystyle{IEEEtran}
\bibliography{library}

\end{document}